\documentclass[a4paper,fleqn]{cas-dc}

\usepackage[authoryear]{natbib}
\usepackage[utf8]{inputenc}
\usepackage{graphicx}
\usepackage{cite}
\usepackage{amsmath,amssymb,amsfonts}
\usepackage{algorithmic}
\usepackage{textcomp}
\usepackage{multirow}
\usepackage{array}

\usepackage[authoryear]{natbib}
\usepackage[utf8]{inputenc}
\usepackage{kotex} 
\usepackage{tabularx}
\usepackage{graphicx}

\newcolumntype{P}[1]{>{\centering\arraybackslash}p{#1}}

\begin{document}
\sloppy
\let\WriteBookmarks\relax
\def\floatpagepagefraction{1}
\def\textpagefraction{.001}
\shorttitle{PCSCNet: Point convolution and Sparse convolution based semantic segmentation}
\shortauthors{Jaehyun Park, Chansoo Kim et~al.}

\title [mode = title]{PCSCNet: Fast 3D Semantic Segmentation of LiDAR Point Cloud for Autonomous Car using Point Convolution and Sparse Convolution Network}                    

\author[1]{Jaehyun Park}[
                        ]
\fnmark[1]

\author[2]{Chansoo Kim}[
                       ]
\fnmark[1]                      
\author[3]{Kichun Jo}[
                      orcid=0000-0003-0543-2198
                      ]
\ead{kichun@konkuk.ac.kr}
\ead[URL]{https://www.auto.konkuk.ac.kr}
\cormark[1]

\address[1]{Department of Automotive Engineering, Hanyang University, Seoul 04763, Republic of Korea}
\address[2]{Department of Intelligent Mobility, Chonnam University, Gwangju 61186, South Korea}
\address[3]{Department of Smart Vehicle Engineering, Konkuk University, Seoul 05029, Republic of Korea}

\cortext[cor1]{Corresponding author}
\fntext[fn1]{These authors contributed equally to this paper.}




\begin{abstract}
The autonomous car must recognize the driving environment quickly for safe driving. As the Light Detection And Range (LiDAR) sensor is widely used in the autonomous car, fast semantic segmentation of LiDAR point cloud, which is the point-wise classification of the point cloud within the sensor framerate, has attracted attention in recognition of the driving environment. Although the voxel and fusion-based semantic segmentation models are the state-of-the-art model in point cloud semantic segmentation recently, their real-time performance suffer from high computational load due to high voxel resolution. In this paper, we propose the fast voxel-based semantic segmentation model using Point Convolution and 3D Sparse Convolution (PCSCNet). The proposed model is designed to outperform at both high and low voxel resolution using point convolution-based feature extraction. Moreover, the proposed model accelerates the feature propagation using 3D sparse convolution after the feature extraction. The experimental results demonstrate that the proposed model outperforms the state-of-the-art real-time models in semantic segmentation of SemanticKITTI and nuScenes, and achieves the real-time performance in LiDAR point cloud inference.
\end{abstract}




\maketitle


\section{Introduction}
\label{sec:introduction}
Fast and accurate recognition of the driving environment is a core concept for an autonomous car. The recognition system must provide the exact position and class information of the objects in real-time. This information is used to determine the regions where the other vehicle and pedestrians do not exist and generate a safe driving path based on the prediction of the dynamic objects’ future motion. 

The recognition system in autonomous car gathers the information of the driving environment using various sensors such as a camera, Radar (Radio detection and ranging), and LiDAR (Light Detection And Ranging). Among those sensors, the LiDAR sensor has some advantages. The LiDAR sensor uses lasers to scan its surroundings and then produces a point cloud. The point cloud provides high-resolution information about the driving environment. It contains an accurate position and precise shape information of the objects in the scanned area. Also, since the point cloud is robust under the illumination change, the LiDAR sensor data are reliable even when the vehicle is in a tunnel or driving at night.

For LiDAR-based recognition, It is essential to quickly extract the semantic information from the LiDAR point cloud using semantic segmentation. The semantic segmentation in LiDAR is to classify the point cloud point-wisely. By semantically segmenting the LiDAR point cloud, the autonomous vehicle can get the accurate location of the objects and also their class information. The point cloud with class information allows the autonomous vehicle to know whether the object is movable and where the drivable space is. 

\begin{figure}[width=.48\textwidth, !t]
\centering
\includegraphics[width=.48\textwidth]{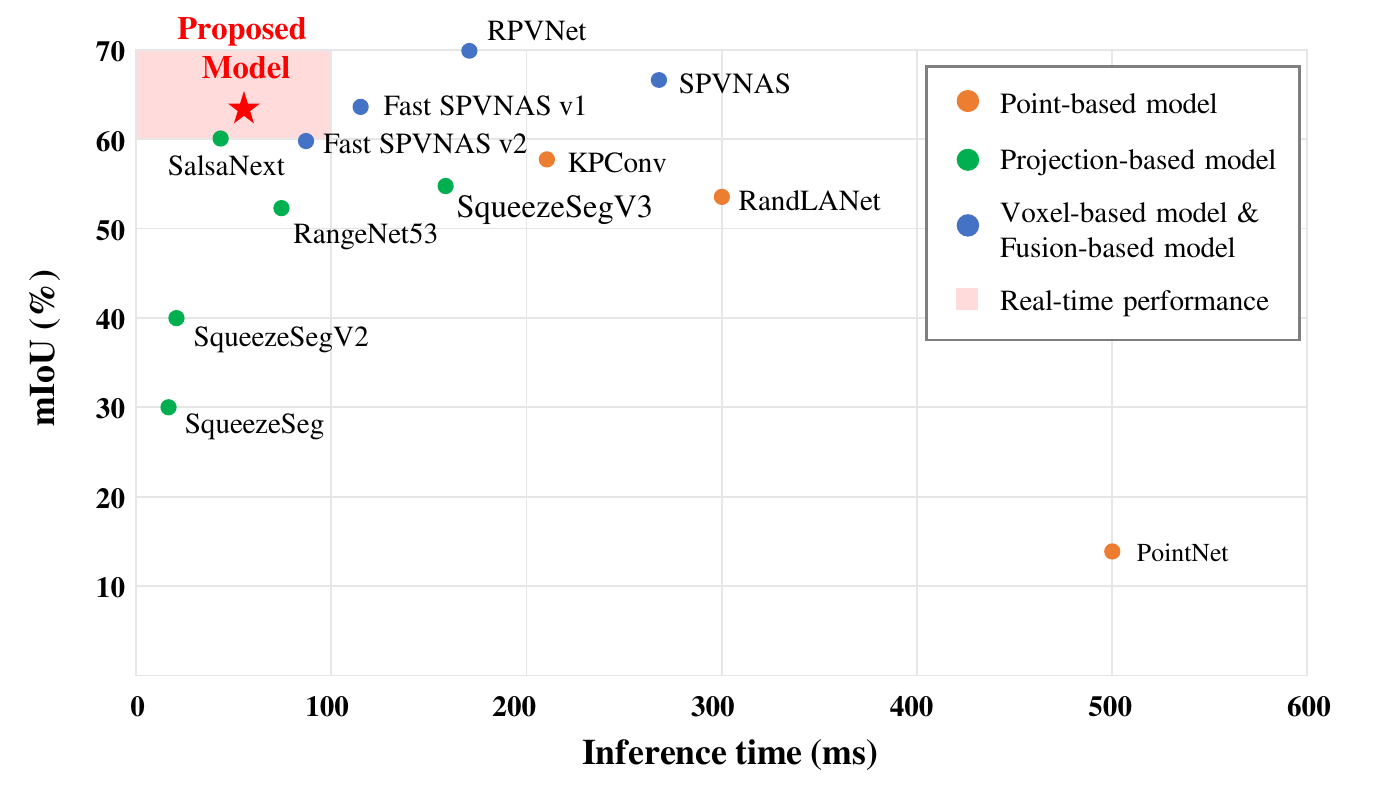}
\caption{Real-time semantic segmentation performance of the proposed model compared to the previous researches. The proposed model runs faster than the recent state-of-the-art models while maintaining the semantic segmentation performance.}
\label{fig:ModelPerformance}
\end{figure}

Under this background, Deep learning-based semantic segmentation of LiDAR point cloud have been researched, and their methods can be categorized in four groups: point-based method, projection-based method, voxel-based method, and fusion-based method. The point-based method directly operates on the input point cloud without any preprocessing. The projection-based method transforms 3D point cloud into 2D image and applied 2D CNN to segment the projected image. On the other hand, the voxel-based method voxelizes the point cloud, and extract the semantic information using 3D CNN. The fusion-based method conducts the voxel-based method and one of the other method in parallel, and fuse the features from each method. As shown in Fig. \ref{fig:ModelPerformance}, the voxel-based method and fusion-based method show the best performance in semantic segmentation, and recently attracted more attention than other methods.

Nevertheless, the voxel-based and fusion-based method have limitation in fast semantic segmentation task due to the computation complexity. The recent voxel-based method compress the multiple points in the voxel to one feature using PointNet \citep{Qi2017a}. Even though PointNet is efficient in applying the deep neural network to point cloud directly, It has weakness in the modeling of long-range contextual information. Therefore, the voxel-based method usually keeps the voxel resolution high (usually 0.05m) to maintain the performance. Thus, those methods inevitably needs a lot of computations in point cloud semantic segmentation.

To overcome this problem, this paper proposes PCSCNet, which is fast voxel-based semantic segmentation model using \textbf{P}oint \textbf{C}onvolution and 3D \textbf{S}parse \textbf{C}onvolution network. For the fast inference speed, PCSCNet voxelizes the point cloud with large size of voxel, and extracts the voxel-wise feature of each voxel using point convolution \citep{Thomas2019} instead of PointNet \citep{Qi2017a}. And then, 3D sparse convolution \citep{Yan2018b} in PCSCNet propagates the voxel-wise features to the neighbor regions to quickly widen the receptive field. Finally, PCSCNet classifies point by point based on the voxel-wise features from point convolution and 3D sparse convolution. 

We summarize the main contributions as:
\begin{itemize}
  \item To reduce the amount of computation in voxel-based semantic segmentation, we designed the point convolution and 3D sparse convolution based semantic segmentation framework called PCSCNet, which has advantage in the low voxel resolution.
  \item To avoid the discretization error from the large-sized voxel, we optimized PCSCNet using the cross-entropy loss and position-aware (PA) loss.
  \item To validate PCSCNet, we conduct quantitative and qualitative experiments, and compare the real-time semantic segmentation performance with the previous methods using two benchmark dataset: SemanticKITTI \citep{Behley2019} and nuScenes \citep{Caesar_2020_CVPR} dataset.
\end{itemize}

The remainder of this paper consists of as follows: Section \ref{section:relatedwork}
 reviews the previous studies about semantic segmentation models. Then, Section \ref{section:Proposed model} and Section \ref{section:loss} present details of the proposed architecture and the network optimization strategy. Section \ref{section:Experiments} gives the experimental results and Section \ref{section:Conclusion} is for the conclusion of this paper

\begin{figure*}[width=\textwidth, !t]
\centering
\includegraphics[width=\textwidth]{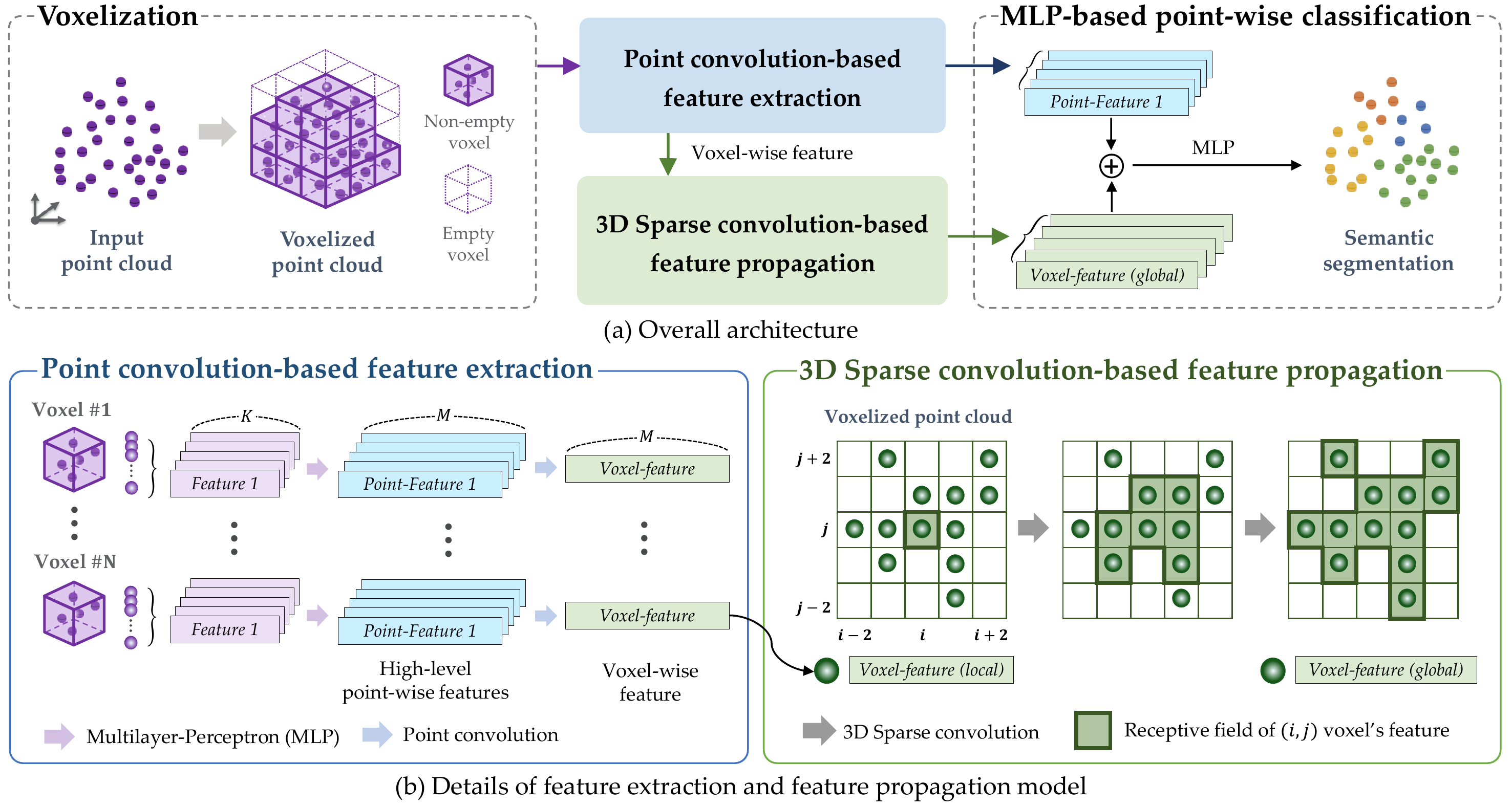}
\caption{Overall architecture of PCSCNet (a) The four steps of the proposed model and the example of the each step's input and output (b) The detail of Point convolution-based feature extraction step and 3D Sparse convolution-based feature propagation step. The purple rectangle is the raw point, the blue rectangle is the high-level point-wise feature from MLP. The green rectangle describes the voxel-wise feature.}
\label{fig:overall}
\end{figure*} 

\section{Related work}
\label{section:relatedwork}
In this section, we briefly explain the previous research about point cloud semantic segmentation. The previous researches can be grouped into 4 groups according to the deep learning method: Point-based method, Projection-based method, Voxel-based method, and Fusion-based method.

\subsection{Point-based method}
PointNet \citep{Qi2017a} is a representative method of point-wise MLP-based semantic segmentation. PointNet proposed a deep learning model using the MLP and max-pooling to extract point-wise features and a global feature of the point set. The proposed architecture successfully extracts the point-wise feature from irregular and unordered point set using MLP. However, PointNet segments the input point cloud with a sliding window for semantic segmentation. Therefore its contextual information is constrained within the size of the window.

Then, the point convolution-based methods \citep{Wang2018a, Hua2018, Hermosilla2018, Thomas2019, Hu2019,  Zhang2019} are proposed to challenge the limitation of the previous point-wise method. Those methods defined the convolution algorithm which can directly apply to the input point cloud without any preprocessing such as 2D projection and voxelization. They extract point-wise features by convolving the point with neighboring points using the correlation function and it accomplished better performance than point-wise MLP-based methods. 

\subsection{Projection-based method}
As 2D CNN demonstrates excellent image segmentation performance \citep{Garcia-Garcia2018}, the 2D CNN-based methods \citep{Wu2018, Wu2018a, Milioto2019, Cortinhal2020, Xu2020, Zhang2020polar, Razani2021} try to apply 2D CNN for the point cloud's semantic segmentation. Those methods project the point cloud into the 2-dimensional image and apply the 2D CNN for the semantic segmentation. Finally, they obtained the semantically segmented point cloud after reconstructing the 2D point cloud into 3D. These methods generally show a fast running time. However, they suffer from distortion of point cloud when the projection occurs and the occlusion between two or more points in the projection of multiple LiDAR point cloud.

\subsection{Voxel-based method}
Initial voxel-based methods \citep{Maturana2015, Riegler2017, Ben-Shabat2018} voxelize the point cloud and apply the 3D CNN for semantic segmentation. Although these methods keep the original dimension of the point cloud, the performance and running time depend on the voxel-resolution. They inevitably have a trade-off relationship between segmentation performance and running time.

As the sparse convolution is proposed in \citep{Graham2014, Yan2018b, Retinskiy2019}, recent voxel-based methods \citep{Yan2020, Tang2020, Zhu2020} accelerates the inference speed by removing the useless computations from empty voxels in point cloud, and also accomplishes successful semantic segmentation performance. However, those models are still not adequate in real-time performance due to the high voxel-resolution.

\subsection{Fusion-based method}
Fusion-based methods \citep{Liong2020, Zhang2020, Xu2021, Cheng2021} use two or more of the previous methods to extract features from point clouds such as combination of voxel and point-based method, or voxel, point, projection-based method. Those methods infer the different backbones in parallel and fuse the point-wise features from each backbone using concatenation, attention, and add. Fusion-based methods are usually state-of-the-art in semantic segmentation. However, as huge amount of computations are required, they are not applicable in real-time task.


\section{PCSCNet}
\label{section:Proposed model}
In this section, we briefly summarize the overall architecture of the proposed PCSCNet and explain the details of the model step-by-step. Our main objective is to accomplish the accurate semantic segmentation of voxelized point cloud while keeping the voxel size largely for fast inference. For the purpose, we designed the voxel-based semantic segmentation model using point convolution \citep{Thomas2019} and 3D sparse convolution \citep{Yan2018b}. 

\subsection{Overall architecture}
PCSCNet can be divided into four steps as shown in Fig. \ref{fig:overall}(a). After voxelization, the point convolution-based feature extraction like the blue box in Fig. \ref{fig:overall}(b) gathers the features from input point cloud. This step extracts the high-dimensional feature of each point (blue rectangles) using MLP. Then, kernel point convolution \citep{Thomas2019} aggregates the high-dimensional point-wise features into one representative voxel-wise feature (green rectangles) of all the voxels.

The 3D sparse convolution-based feature propagation, described in the green box in Fig. \ref{fig:overall}(b), takes the voxel-wise features from the previous step as inputs. Then, this step fastly propagates the input features to the neighboring voxels using 3D sparse convolution. The feature propagation step outputs the global voxel-wise features which contain the wide range information. 

The final step is the MLP-based point-wise classification. As shown in Fig. \ref{fig:overall}(a), this step concatenates the point-wise features and voxel-wise features from the previous two steps, and then classifies each point by applying the MLP to the concatenated features. The details of each step are described in the following sections.

\subsection{Point convolution-based feature extraction}
The raw point cloud from the LiDAR sensor is the set of unordered points. To apply the CNN for semantic segmentation of point cloud, we first convert the point cloud into the regular data format using voxelization. The voxelization represents a point cloud with some 3D grid cells called voxel and indices each point with its corresponding voxel. However, each voxel in the voxelized point cloud contains only simple information such as the center coordinate and the point’s position in its range. This information is not enough for semantic segmentation of the point cloud. Therefore, PCSCNet obtains more details of each voxel using the point convolution based feature extraction model inspired by kernel-point convolution \citep{Thomas2019}.

\begin{figure}[width=.48\textwidth, t]
\includegraphics[width=.48\textwidth]{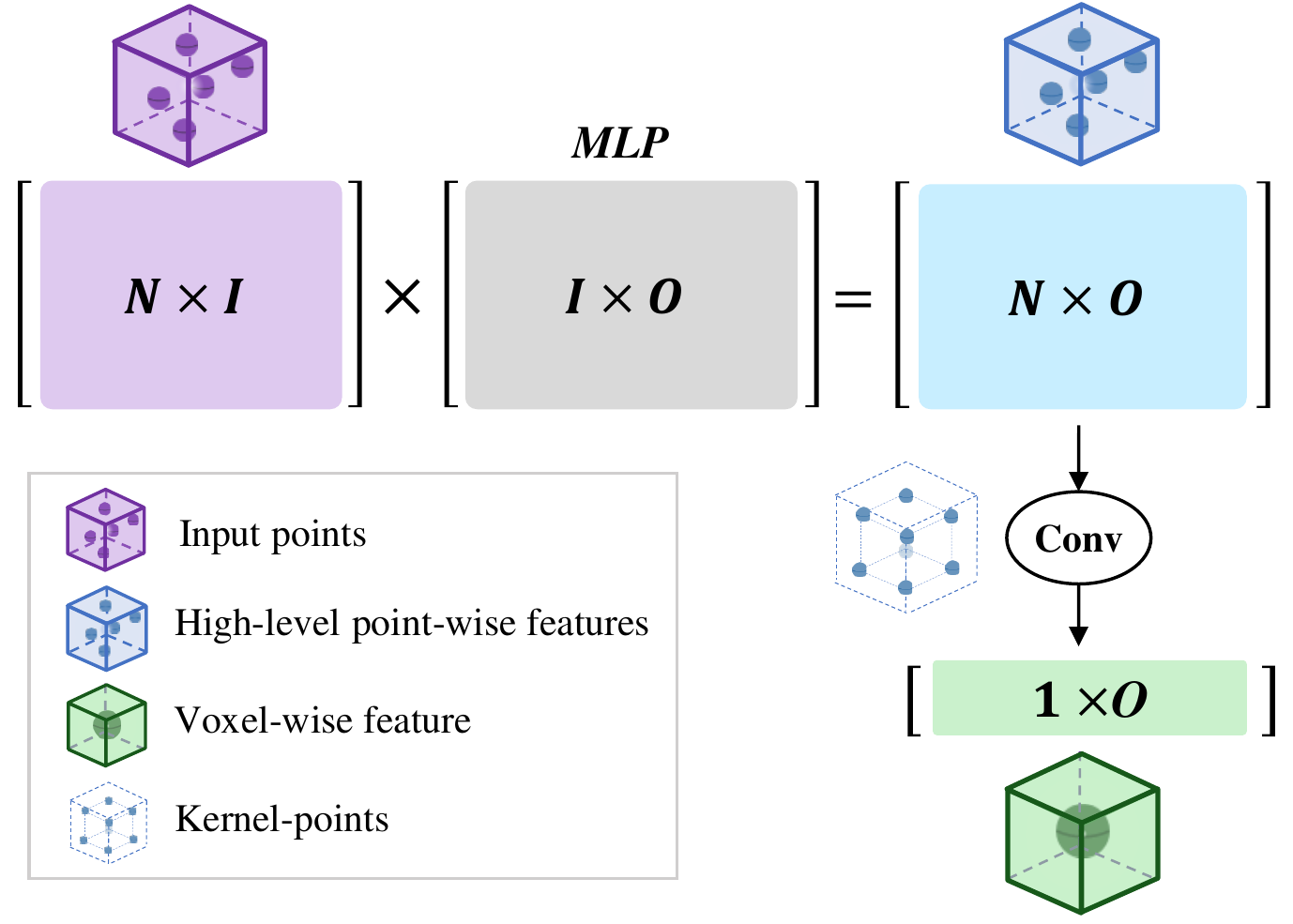}
\caption{The extraction of voxel-wise feature(green) and high-level point-wise features(blue) from the points(purple) using MLP (gray) and point-convolution}
\label{fig:pointnet}
\end{figure}

Kernel-point convolution \citep{Thomas2019} proposed the continuous convolution which can be directly applied to the point set in point cloud. The details of kernel-point convolution-based voxel-wise feature extraction model are as follow. For the clarity, Let \({x_{c}}\in\mathbb{R}^3\) is the center position of the voxel, and the voxel contains the \(N\) points like a purple voxel in Fig. \ref{fig:pointnet}. \(\mathcal{X}\in\mathbb{R}^{N\times3}\) is the coordinates of the corresponding points of the voxel and \(\mathcal{P}\in\mathbb{R}^{N\times{I}}\) is the corresponding input features. The voxel-wise feature extraction model output of this voxel is \(v\in\mathbb{R}^{1\times{O}}\), which is described as a green voxel in Fig. \ref{fig:pointnet}.

The first step is to obtain the high-level point-wise features \(\mathcal{F}\in\mathbb{R}^{N\times{O}}\), which are blue rectangle in Fig. \ref{fig:pointnet}, from the input feature set \(\mathcal{P}\) using MLP. The \(i^{th}\) input feature (\(p_i\)) is expanded to the \(O\) dimensional feature (\(f_i\)) after MLP layer.

\begin{equation}
{f_i}=\text{ }MLP({p_{i}}), \text{where } MLP:\mathbb{R}^I\rightarrow\mathbb{R}^O
\label{eqn:mlp}
\end{equation}

Then, those high-level features \(\mathcal{F}\) aggregate to the voxel-wise feature \(v\) through the convolution with kernel-point set. We call the number kernel points as \(K\), and its position as \(\mathcal{K}\in\mathbb{R}^{K\times{3}}\), and one of its convolution weights as  \(\mathcal{W}_k\in\mathbb{R}^{O\times{O}}\). Based on the convolution in image, we can define the kernel-point convolution proposed in KPConv \citep{Thomas2019} like:

\begin{equation}
v=\sum_{k\in\mathcal{K}}\Big\{\sum_{x_i\in\mathcal{X}}{h(x_i-x_c, k)f_iW_k}\Big\}
\label{eqn:kpconv}
\end{equation}

The function \(h\) in the equation calculates the correlation between the corresponding points and each kernel point. In case of image convolution, it is defined in discrete domain and it is easy to relate between the pixel and kernel based on their coordinates. On the other hand, the point convolution is in continuous domain. It is necessary to define the correlation function between input point and kernel-point, which is increasing its output when the kernel-point and the input point are being close. As proposed in KPConv \citep{Thomas2019}, we use the correlation function as follows.

\begin{equation}
h(x, k)=max\left(0, 1-\frac{\parallel{x-k}\parallel}{\sigma}\right)
\label{eqn:correleation}
\end{equation}

The \(\sigma\) determines how fast the correlation value is decreasing when the distance between the kernel point and input point is far. In the proposed feature extraction model, we decide the \(\sigma\) depends on the number of kernel points and the size of voxel. 

\subsection{3D Sparse convolution-based feature propagation}
The voxel-wise features from the Point convolution-based feature extraction step only contain the information of each voxel. For the long-range contextual information of the voxel-wise feature, PCSCNet propagates the voxel-wise features to the neighbors using 3D sparse convolution which can accelerates the computation at the sparse input.

\subsubsection{Regular convolution and Sparse convolution}
Both Regular convolution \citep{Ciregan2012} and Sparse convolution \citep{Yan2018b}  propagates and aggregates the features of the geometrically neighboring regions. From those convolution layers, we can extract the simple pattern such as vertical or horizontal shape in the input data. Although both convolution targets to expand the local features to global features, the convolution mechanisms are totally different in a sparse data like handwriting image.

\begin{figure}[width=.48\textwidth, t]
\includegraphics[width=.48\textwidth]{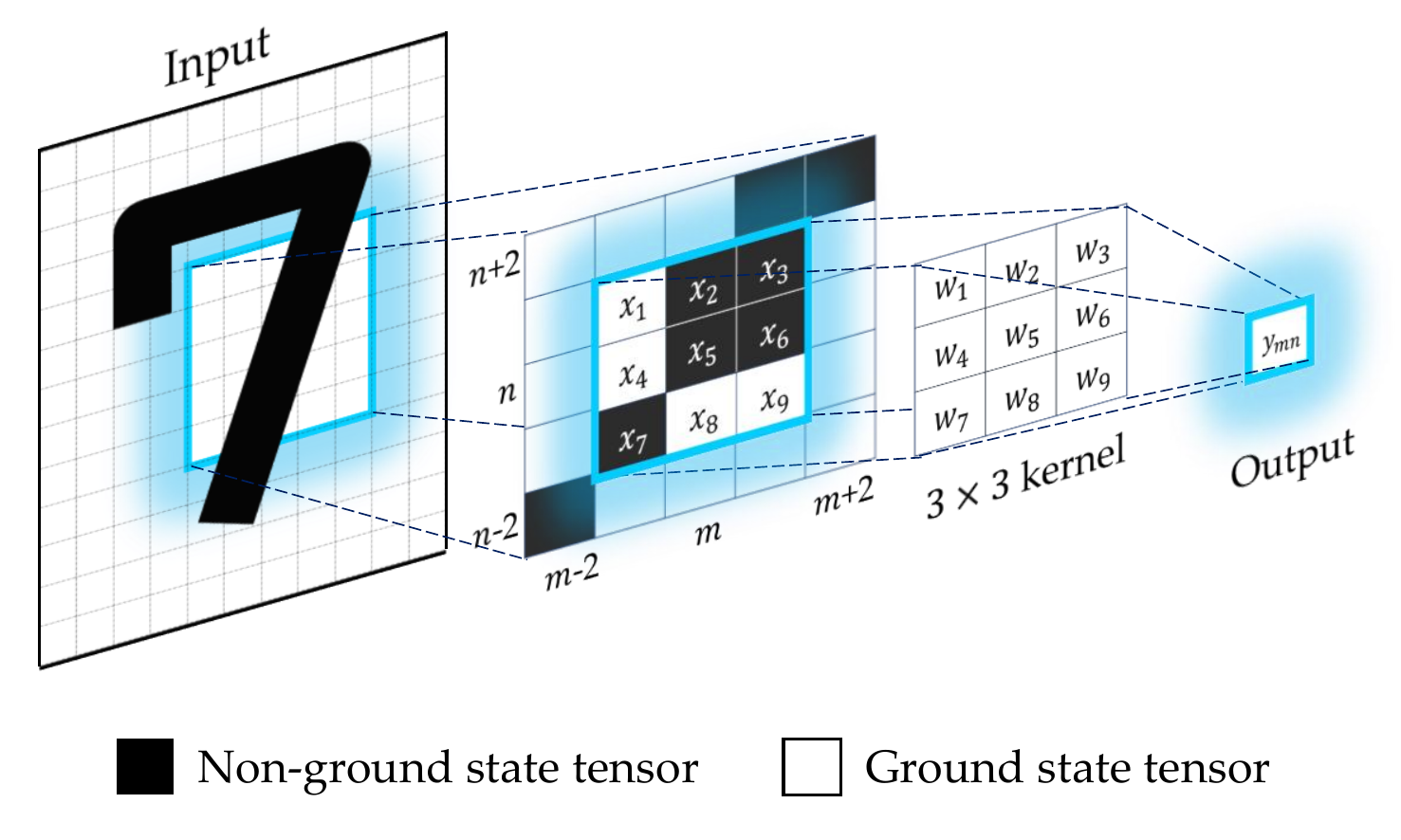}
\caption{Example of 3 by 3 convolution at (3, 3) pixel in sparse data(handwriting image). The sparse data consists of the non-ground state tensors(black) and ground state tensors(white). The non-ground state tensors contain the meaningful information of the handwriting image unlike the ground tensors.}
\begin{multline*}
y_{mn}=\sum_{i = 1}^{9}\sum_{j = 1}^{9}{x_{i}w_{j}1(i, j)},\\
\text{where }1(i, j)= 
\begin{cases}
    1,              & \text{if } i=j\\
    0,              & \text{otherwise} 
\end{cases}
\end{multline*}

\begin{multline*}
y_{mn}=\sum_{i = 1}^{9}\sum_{j = 1}^{9}{x_{i}w_{j}g(i, j)}, \\ 
\text{where }g(i, j)=
\begin{cases}
    1,              & \text{if } i=j \text{ and}\\
                    & x_{i}\text{ is a }\emph{non ground state}\\
    0,              & \text{otherwise}
\end{cases}
\label{eqn:sparse}
\end{multline*}
\label{fig:hand}
\end{figure}

Regular convolution \citep{Ciregan2012} accomplshed great success in image classification and segmentation task. By convolving the convolutional kernel with input, It efficiently extracts the features with less parameters while keeping the geometry of input. However, the efficiency of Regular convolution is valid only in dense input and fixed size of input. The regular convolution algorithm is not applicable in sparse data. Fig. \ref{fig:hand} is a handwriting image which is the example of sparse data. As shown in Fig. \ref{fig:hand}, the sparse data has \emph{ground-state} tensors and \emph{non-ground-state} tensors. The \emph{ground-state} tensors are not meaningful regions like white pixels in the handwriting image. Otherwise, the \emph{non-ground-state} tensors contain the information of the input data like black pixels. Fig. \ref{fig:hand} visuliazes the 3 by 3 convolution in the \((m,n)\) pixel. \(x_i\) represents the subset of the input tensors which corresponds to the 3 by 3 convolutional kernel and \(w_i\) is the weight of kernel. As the result of the convolution, the input tensors aggregate to one output tensor \(y_m{}_n\). As described in the upperline of the above equation, the convolutional kernel in Regular convolution takes all the corresponding regions as an input, even though those regions include the \emph{ground-state} tensors. As computing the \emph{ground-state} tensors is useless and a waste of memory resources, Regular convolution is inefficient in sparse data.

Unlike Regular convolution, the Sparse convolution \citep{Yan2018b} takes only the \emph{non-ground-state} tensors as an input when convolving the sparse data like the underline of the above equation. The sparse convolution judges whether the corresponding tensors are in the \emph{ground-state} or not and convolves only the \emph{non-ground-state} tensors. Then, Sparse convolution saves more computing resources than Regular convolution. Also, Sparse convolution is memory efficient because it consumes the memory only for the \emph{non-ground-state} tensors.

\begin{figure}[width=.48\textwidth, !h]
\includegraphics[width=.48\textwidth]{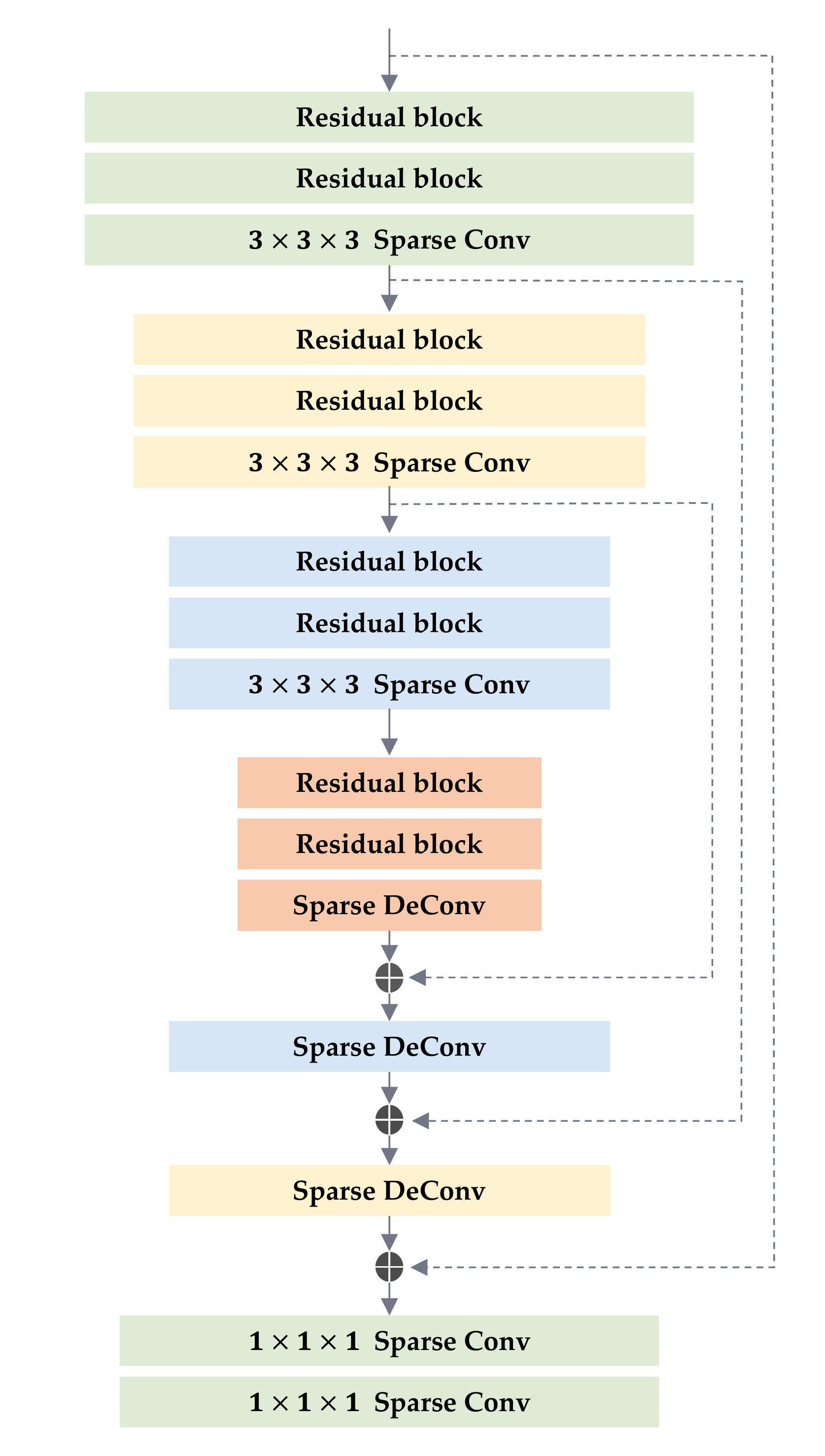}
\caption{3D Sparse convolution based UNet architecture to propagate the voxel-wise features globally (Residual block, Sparse-conv block), and to reconstruct the dimension of original voxelized point cloud (Sparse-deconv block).}
\label{fig:arch}
\end{figure} 

\subsubsection{3D Sparse convolution-based voxel-wise feature propagation}
The voxelized point cloud is also the sparse data that contains non-empty voxels and empty voxels. The non-empty voxel has the points within its range and it is regarded as the \emph{non-ground-state} tensor. On the other hand, the empty voxel is the \emph{ground-state} tensor as it does not have any points. Therefore, the proposed model efficiently widens the receptive field of voxel-wise features with 3D Sparse convolution \citep{Yan2018b}.

The architecture of the proposed feature propagation is inspired by U-Net architecture \citep{Ronneberger2015} like Fig. \ref{fig:arch}. It takes the voxel-wise features from the feature extraction model as an input and propagates the features with three sparse convolution blocks. The output of the deeper convolution block covers more extended range so that it is more global information. Finally, the propagation model concatenates each output and up-samples them to reconstruct the original size of the voxelized point cloud. Based on the proposed architecture, the voxels contain the longer-range contextual information than the input voxels have.


\subsection{MLP based point-wise classification}
To accelerate the inference speed of the voxel-based methods, we increase the size of the voxel and it inevitably causes the the discretization error in the boundary points between two voxels. In order to overcome the problem, PCSCNet classifies the point cloud point-wisely in this step. This step uses the features from the previous two steps: the high-level point-wise features from the feature extraction step and the global voxel-wise features from the feature propagation step. 

\begin{figure}[width=.48\textwidth, h]
\includegraphics[width=.48\textwidth]{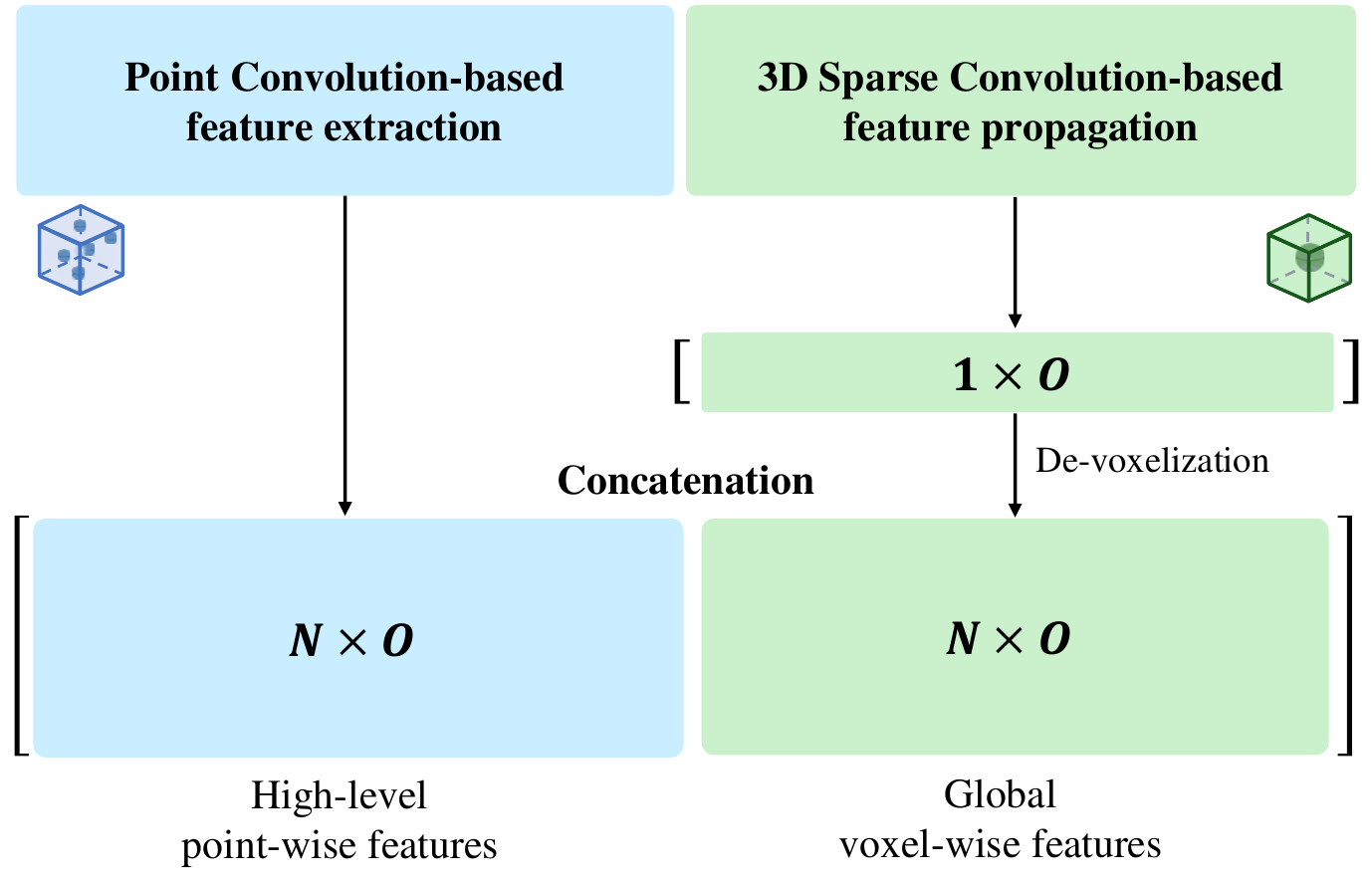}
\caption{Concatenation of the point-wise features and a voxel-wise feature. The blue rectangle is the high-level point-wise features from the point convolution-based feature extraction step. The green rectangle is the voxel-wise features, which contains long-range contextual information, from the 3D sparse convolution-based feature propagation step.}
\label{fig:MLP}
\end{figure}

This step is described in Fig. \ref{fig:MLP}. In this step, the voxel-wise features from the feature propagation model are devoxelized first. The voxel-wise features are copied as many as the number of points in its range, and then they concatenated with the high-level point-wise features. As a result, the concatenated feature has not only the abundant point-wise information and also the global information. Finally, from the concatenated features, MLP layer calculates the probability per class of each point and classifies each point by taking the max value among the class probabilities.

\section{Network optimization}
\label{section:loss}
As we configure the size of voxel largely to accelerate the running time of PCSCNet, the miss-segmentation in boundary points between two different object occurs more frequently than the previous voxel-based methods. In order to overcome this limitation, we optimize the proposed model using the combined loss(\(L\)), which is the weighted sum of cross-entropy loss(\(L_{ce}\)) and position-aware loss(\(L_{pa}\)) like:

\begin{equation}
L(\bold{y}, \bold{\hat{y}})=w_{ce}L_{ce}(\bold{y}, \bold{\hat{y}}) + w_{pa}L_{pa}(\bold{y}, \bold{\hat{y}})
\label{eqn:wce1}
\end{equation}

Position-aware \citep{Liu2020} loss penalties when the model incorrectly predicts the segments at boundary, and it is firstly proposed in image semantic segmentation task. Therefore, the previous Position-aware loss defined in discrete domain. However, point cloud is in continuous domain unlike image, and we propose transformed position-aware loss, which is suitable in point cloud data.

Firstly, we select the boundary points from trained point cloud. In order to pick the boundary points, we compare the label of all the point with their 10 nearest points. Then, if the point has one or more differently labeled points (\(N_{diff}\)) among 10 nearest points, we weight the cross-entropy loss of the point by the number of differently labeled points:

\begin{align}
L_{ce}(\bold{y}, \bold{\hat{y}})=-\sum_{i = 1}^{C}\hat{y}_ilog(y_i) \\
L_{pa}(\bold{y}, \bold{\hat{y}})=N_{diff}L_{ce}(\bold{y}, \bold{\hat{y}})
\label{eqn:wce}
\end{align}

where \(C\) is the number of classes, \(\bold{\hat{y}}\in\mathbb{R}^{1\times{C}}\) is the one-hotted ground truth and \(\bold{y}\in\mathbb{R}^{1\times{C}}\) is the softmax output of the semantic segmentation model. \(y_i\) and \(\hat{y_i}\) are predicted values and the ground truth of \(i\) th class.



\section{Experiments}
\label{section:Experiments}
In this section, we share the experimental results to evaluate our model. The experiments are designed to confirm our model's advantage using two LiDAR semantic segmentation dataset: SemanticKITTI \citep{Behley2019} and nuScenes \citep{Caesar_2020_CVPR}. The semantic segmentation performance and real-time performance of the proposed model is compared with the point-based models, projection-based models and the voxel-based models. Also, we only selected the comparison models where the running time was less than 100 ms.

Section \ref{section:Dataset} briefly explains the experimental datasets. Section \ref{section:implementation} contains the implementation details. Section \ref{section:SemanticKITTI} and Section \ref{section:Nuscenes} provides the experimental results of SemanticKITTI \citep{Behley2019} and nuScenes \citep{Caesar_2020_CVPR}. Section \ref{section:Ablation} discuss the advantages of the proposed approach with ablation studies.

\subsection{Dataset}
\label{section:Dataset}
\subsubsection{SemanticKITTI}
SemanticKITTI Odometry Benchmark \citep{Behley2019} contains over 40,000 point clouds gathered from 22 different sequences using one Velodyne HDL-64E LiDAR. Although each point cloud is point-wisely annotated with 22 classes, we only used 19 classes for training and testing. Among the driving sequences, 10 sequences (00 to 07, 09 to 10) were used to train our model, and sequence 08 was for the hyperparameter selection. The remaining sequences were used for the comparison of our model's performance with previous methods.

\renewcommand{\arraystretch}{1.6}
\begin{table*}[width=\textwidth, h!]
\scriptsize
{

\centering
\begin{tabular}{|@{}P{1.1cm}@{}|@{}P{2.5cm}@{}|@{}P{0.6cm}@{}|@{}P{0.65cm} @{}P{0.65cm} @{}P{0.65cm} @{}P{0.65cm} @{}P{0.65cm} @{}P{0.65cm} @{}P{0.65cm} @{}P{0.65cm} @{}P{0.65cm} @{}P{0.65cm} @{}P{0.65cm} @{}P{0.65cm} @{}P{0.65cm} @{}P{0.65cm} @{}P{0.65cm} @{}P{0.65cm} @{}P{0.65cm} @{}P{0.65cm} @{}P{0.65cm}@{} |@{}P{0.7cm}@{}|}
 \hline
 &&& \multicolumn{19}{c|}{\textbf{IoU per Class (\%)}} & \\
 \cline{4-22}
                      \rotatebox{90}{\textbf{Methods}}
                      & \textbf{Network}
                      & \rotatebox{90}{\textbf{Scan per Sec}}
                      & \rotatebox{90}{car} 
                      & \rotatebox{90}{bicycle} 
                      & \rotatebox{90}{motorcycle} 
                      & \rotatebox{90}{truck} 
                      & \rotatebox{90}{other-vehicle} 
                      & \rotatebox{90}{person} 
                      & \rotatebox{90}{bicyclist} 
                      & \rotatebox{90}{motorcyclist} 
                      & \rotatebox{90}{road} 
                      & \rotatebox{90}{parking} 
                      & \rotatebox{90}{sidewalk} 
                      & \rotatebox{90}{other-ground} 
                      & \rotatebox{90}{building} 
                      & \rotatebox{90}{fence} 
                      & \rotatebox{90}{vegetation} 
                      & \rotatebox{90}{trunk} 
                      & \rotatebox{90}{terrain} 
                      & \rotatebox{90}{pole} 
                      & \rotatebox{90}{traffic-sign} 
                      & \rotatebox{90}{\textbf{mean IoU (\%)}}\\
  \hline\hline
  \multirow{3}{*}{\rotatebox{90}{Point}} &PointNet & 0.2                                         &46.3&1.3&0.3&0.1&0.8&0.2&0.2&0.0&61.6&15.8&35.7&1.4&41.4&12.9&31.0&4.6&17.6&2.4&3.7&14.6 \\
                          &RandLANet&0.3&94.2&26.0&25.8&40.1&38.9&49.2&48.2&7.2&90.7&60.3&73.7&20.4&86.9&56.3&81.4&61.3&66.8&49.2&47.7&53.9 \\
                         &KPConv&0.4&96.0&30.2&42.5&33.4&44.3&61.5&61.6&11.8&88.8&61.3&72.7&31.6&90.5&64.2&84.8&69.2&69.1&56.4&47.4&58.8 \\
 \hline\hline
 \multirow{6}{*}{\rotatebox{90}{Projection}} 
 &SqueezeSeg &27&68.8&16.0&4.1&3.3&3.6&12.9&13.1&0.9&85.4&26.9&54.3&4.5&57.4&29.0&60.0&24.3&53.7&17.5&24.5&29.5\\
 &SqueezeSegV2 &24&81.8&18.5&17.9&13.4&14&20.1&25.1&3.9&88.6&45.8&67.6&17.7&73.7&41.1&71.8&35.8&60.2&20.2&36.3&39.7\\
 &RangeNet++ &16&91.4&25.7&34.4&25.7&23.0&38.3&38.8&4.8&91.8&65.0&75.2&27.8&87.4&58.6&80.5&55.1&64.6&47.9&55.9&52.2\\
 &PolarNet &20&83.8&40.3&30.1&22.9&28.5&43.2&40.2&5.6&90.8&61.7&74.4&21.7&90.0&61.3&84.0&65.5&67.8&51.8&57.5&54.3\\
 &SalsaNext &24&91.9&48.3&38.6&38.9&31.9&\textbf{60.2}&59.0&19.4&91.7&63.7&75.8&29.1&90.2&64.2&81.8&63.6&66.5&54.3&62.1&59.5\\
 &Lite-HDSeg &20&92.3&40.0&\textbf{55.4}&37.7&39.6&59.2&\textbf{71.6}&54.1&93.0&68.2&\textbf{78.3}&29.3&91.5&65.0&78.2&65.8&65.1&59.5&\textbf{67.7}&63.8\\
 \hline\hline
 \multirow{3}{*}{\rotatebox{90}{\shortstack{Voxel\\\& Fusion}}} 
 &{SPVNAS}&11&-&-&-&-&-&-&-&-&-&-&-&-&-&-&-&-&-&-&-&60.3 \\
 &FusionNet&11&95.3&47.5&37.7&\textbf{41.8}&34.5&59.5&56.8&11.9&91.8&\textbf{68.8}&77.1&\textbf{30.8}&\textbf{92.5}&\textbf{69.4}&\textbf{84.5}&\textbf{69.8}&\textbf{68.5}&60.4&66.5&61.3 \\
 \cline{2-23}
 &\textbf{PCSCNet (Ours)}&16&\textbf{95.7}&\textbf{48.8}&46.2&36.4&\textbf{40.6}&55.5&68.4&\textbf{55.9}&89.1&60.2&72.4&23.7&89.3&64.3&84.2&68.2&68.1&\textbf{60.5}&63.9&62.7\\
 \hline
\end{tabular}
\caption{Comparison of latency, frames per second, and Intersection of Union(IoU) per class with previous models using the test sequences(11 to 21) in SemanticKITTI.}
\label{table:kitti}
}
\end{table*}

\begin{table*}[width=\textwidth, h!]
\scriptsize
{
\centering
\begin{tabular}{|P{2.5cm}|@{}P{0.8cm}@{} @{}P{0.8cm}@{} @{}P{0.8cm}@{} @{}P{0.8cm}@{} @{}P{0.8cm}@{} @{}P{0.8cm}@{} @{}P{0.8cm}@{} @{}P{0.8cm}@{} @{}P{0.8cm}@{} @{}P{0.8cm}@{} @{}P{0.8cm}@{} @{}P{0.8cm}@{} @{}P{0.8cm}@{} @{}P{0.8cm}@{} @{}P{0.8cm}@{} @{}P{0.8cm}@{}|@{}P{0.8cm}@{}|}
 \hline
 & \multicolumn{16}{c|}{\textbf{IoU per Class (\%)}} & \\
 \cline{2-17}
                      {\textbf{Network}} 
                      & \rotatebox{90}{barrier} 
                      & \rotatebox{90}{bicycle} 
                      & \rotatebox{90}{bus} 
                      & \rotatebox{90}{car} 
                      & \rotatebox{90}{construction} 
                      & \rotatebox{90}{motorcycle} 
                      & \rotatebox{90}{pedestrian} 
                      & \rotatebox{90}{traffic-cone} 
                      & \rotatebox{90}{trailer} 
                      & \rotatebox{90}{truck} 
                      & \rotatebox{90}{drivable} 
                      & \rotatebox{90}{other-flat} 
                      & \rotatebox{90}{sidewalk} 
                      & \rotatebox{90}{terrain} 
                      & \rotatebox{90}{manmade} 
                      & \rotatebox{90}{vegetation} 
                      & \rotatebox{90}{\textbf{mean IoU (\%)}}\\
  \hline\hline
               RangeNet++  & 66.0 & 21.3 & 77.2 & 80.9 & 30.2 & 66.8 & 69.6 & 52.1 & 54.2 & 72.3 & 94.1 & 66.6 & 63.5 & 70.1 & 83.1 & 79.8 & 65.5 \\
               PolarNet & 74.7 & 28.8 & 85.3 & \textbf{90.9} & 35.1 & 77.5 & 71.3 & 58.8 & 57.4 & 76.1 & \textbf{96.5} & \textbf{71.1} & \textbf{74.7} & \textbf{74.0} & \textbf{87.3} & \textbf{85.7} & 71.0 \\
               SalsaNext & \textbf{74.8} & 34.1 & 85.9 & 88.4 & 42.2& 72.4 & 72.2 & \textbf{63.1}&  \textbf{61.3} & 76.5 & 96.0 & 70.8 & 71.2 & 71.5 & 86.7 & 84.4 & 71.9 \\
   \hline
               \textbf{PCSCNet (Ours)} & 73.3 & \textbf{42.2} & \textbf{87.8} & 86.1 & \textbf{44.9} & \textbf{82.2} & \textbf{76.1} & 62.9 & 49.3 & \textbf{77.3} & 95.2 & 66.9 & 69.5 & 72.3 & 83.7 & 82.5 & \textbf{72.0} \\

 \hline
\end{tabular}
\caption{Comparison of Intersection of Union(IoU) per class with previous models using the validation data in nuScenes lidarSeg}
\label{tab:tabnu}
}
\end{table*}

\begin{figure*}[width=\textwidth, h!]
\centering
\includegraphics[width=\textwidth]{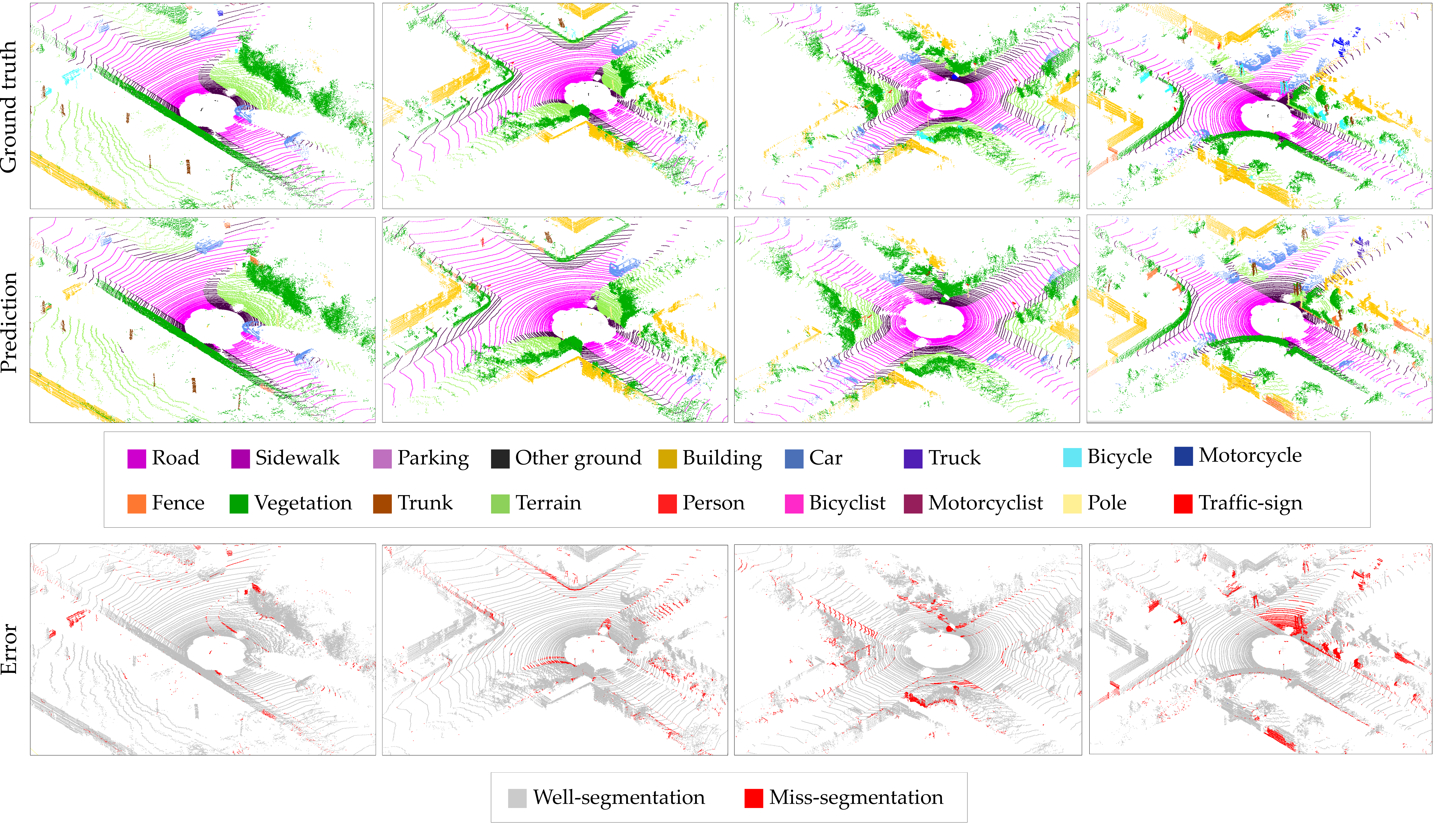}
\caption{Visualizing the semantic segmentation results of the proposed model using 08 sequence in SemanticKITTI. The first row is the ground truth, the second row is the prediction of the proposed model, and the third row is the difference between the ground truth and prediction. We observe that the proposed model segments well in usual road scenario under this experiment.}
\label{fig:figsemantic}
\end{figure*} 

\subsubsection{nuScenes}
nuScenes LidarSeg dataset \citep{Caesar_2020_CVPR} also provides about 40,000 labeled point clouds, which were collected from both Singapore and Boston using Velodyne HDL-32E. The nuScenes dataset labeled with 31 classes. Among them, 23 classes are foreground classes and the others are background stuff. In this paper, those 31 classes are mapped into 16 classes to compare with the previous models. Additionally, out of total 40,000 point clouds, 28,130 frames are used for training, 6,019 frames are used for validation, and the remainder for testing.

\subsection{Implementation details}
\label{section:implementation}
Those experiments are conducted on a NVIDIA RTX3090 GPU. The voxel is a cube shape and its size is 0.1m which is bigger than other previous voxel-based methods. At the training step, the proposed model optimized using the Adams optimizer with default parameter setting. Also, \(w_{ce}\) and \(w_{pa}\) in combined loss function are set to 1.0 and 1.5., and . The maximum epoch is set as 150 and initial learning rate is 0.001.

\subsection{Evaluation on SemanticKITTI}
\label{section:SemanticKITTI}
Table \ref{table:kitti} contains a quantitative result of the experiment on SemanticKITTI \citep{Behley2019}. The compared models are the state-of-the-art approaches in fast semantic segmentation task. As described in the table, PCSCNet can infer 16 scans per second. This result demonstrates that our model is as fast as projection-based approaches and is much faster than previous voxel-based and fusion-based approaches. Furthermore, our model achieves the accurate semantic segmentation result in SemanticKITTI. The mIoU (mean Intersection of Union) of our model is the highest among the fast voxel-based methods. Especially, our model significantly outperforms in the segmentation of dynamic objects such as a car, bicycle, motorcycle, and other-vehicle classes. 

Fig. \ref{fig:figsemantic} visualizes the semantic segmentation results of PCSCNet. The test sequence is 08 sequence in SemanticKITTI \citep{Behley2019}. The first column is the visualization of ground truth, and the second column is our predictions. The last column indicates miss-classified points with red color. As you can see, our model well classifies car, terrain, building, and road classes. Especially, the points from car class are segmented well with our model. However, our model has difficulties in the segmentation of the sidewalk, parking-lot points. This result is due to the insufficiency of the points from those classes. As the shape of the parking-lot and sidewalk are similar to the road, the points from the parking-lot and sidewalk are classified into the classes that have much more amount of data such as road.

\subsection{Evaluation on Nuscenes}
\label{section:Nuscenes}
Table \ref{tab:tabnu} is the semantic segmentation results of PCSCNet and the previous models using nuScenes validation data \citep{Caesar_2020_CVPR}. Although the compared models are also the state-of-the-art models in the semantic segmentation task, PCSCNet shows the highest mean IoU in the experiment. As shown in the experiment using nuScenes, the proposed model usually outperforms in the semantic segmentation of dynamic objects such as bicycle, motorcycle, truck and pedestrian. 

Although nuScenes dataset gathered the point cloud using 32 layer LiDAR and it is more sparse than SemanticKITTI, the mIoU of PCSCNet is still higher than other real-time models in this experiment. This result means that our model validates not only in dense point cloud, but also in sparse LiDAR point cloud data. 

\subsection{Ablation studies}
\label{section:Ablation}
We conducted the additional ablation studies to validate the advantages of PCSCNet. In this section, we firstly discuss the robustness of our model under the change of voxel-resolution, and the effect of Position-Aware (PA) loss by comparing with cross-entropy loss.

\subsubsection{Effect of the voxel resolution}
To validate the robustness of PCSCNet under the change of voxel-resolution, we compared the performance of three different models while increasing the voxel size using 08 sequence in SemanticKITTI. One is the voxel-based model, Another model fuses the voxel-based method and point-based method. The other model is PCSCNet.

\begin{figure}[width=.48\textwidth, !h]
\centering
\includegraphics[width=.48\textwidth]{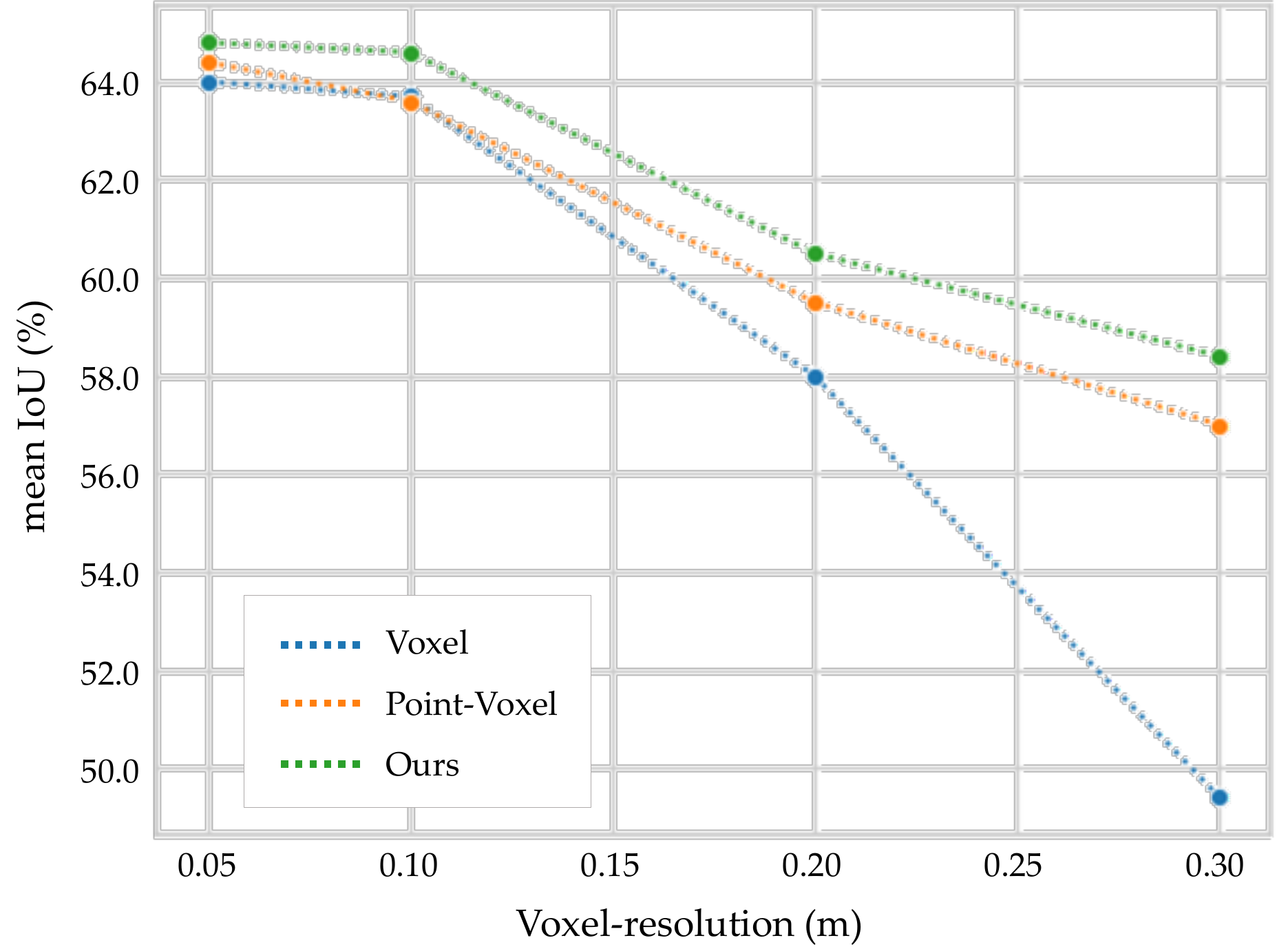}
\caption{The mIoU change under the various voxel resolution: voxel-based model (blue line), point-voxel fusion based model(Orange line), and the proposed model (point convolution and voxel fusion)(Green line).}
\label{fig:VoxRes}
\end{figure}

Fig. \ref{fig:VoxRes} and Table \ref{table:voxres} is the result of the comparison. PCSCNet is about 1\% higher than point-voxel fusion model over the entire range of voxel-resolution. Furthermore, while the point-voxel fusion model sharply degrades in the semantic segmentation performance in 0.05m to 0.1m range, the mIoU of the proposed model drops only 0.2\% in the same range. This result means that the proposed model not only has good performance in high-resolution voxel, but is also less affected by changes in voxel resolution.


    

\begin{table}[!h]
\scriptsize
{
\centering
\begin{tabular}{ |c|c|c|c|}
 \hline
 Voxel & \multicolumn{3}{c|}{mIoU of approaches} \\
 \cline{2-4}
 resolution(m) & Voxel & Point-Voxel & PCSCNet \\
 \hline\hline
      0.05 & 64.0 & 64.4 & \textbf{64.8}\\
      0.10 & 63.7 & 63.6 & \textbf{64.6}\\
      0.20 & 60.0 & 59.5 & \textbf{60.5} \\
      0.30 & 49.4 & 57.0 & \textbf{58.4}\\
     \hline
\end{tabular}
\caption{The mIoU of voxel, point-voxel fusion, and the proposed methods. Those models are trained and tested under the various voxel-size (0.05m to 0.3m) using SemanticKITTI.}
\label{table:voxres}
}
\end{table}

\subsubsection{Effect of the Position-Aware (PA) loss}
We analyzed the effect of the cross-entropy and position-aware combined loss. In the experiment, one PCSCNet is trained with only cross-entropy loss and the other PCSCNet is optimized with weighted sum of cross-entropy and position-aware loss. We trained those models using SemanticKITTI training set and compare the quantitative and qualitative performance using the 08 sequence in SemanticKITTI dataset. 

The quantitative result is shown in Table \ref{table:combinedLoss}. The comparison table demonstrates that the the model trained with the combined loss outperforms the other model in terms of mean IoU. The combined loss trained model is about 0.8\% better than the other model in mIoU. 

\begin{figure}[width=.48\textwidth, !h]
\centering
\includegraphics[width=.48\textwidth]{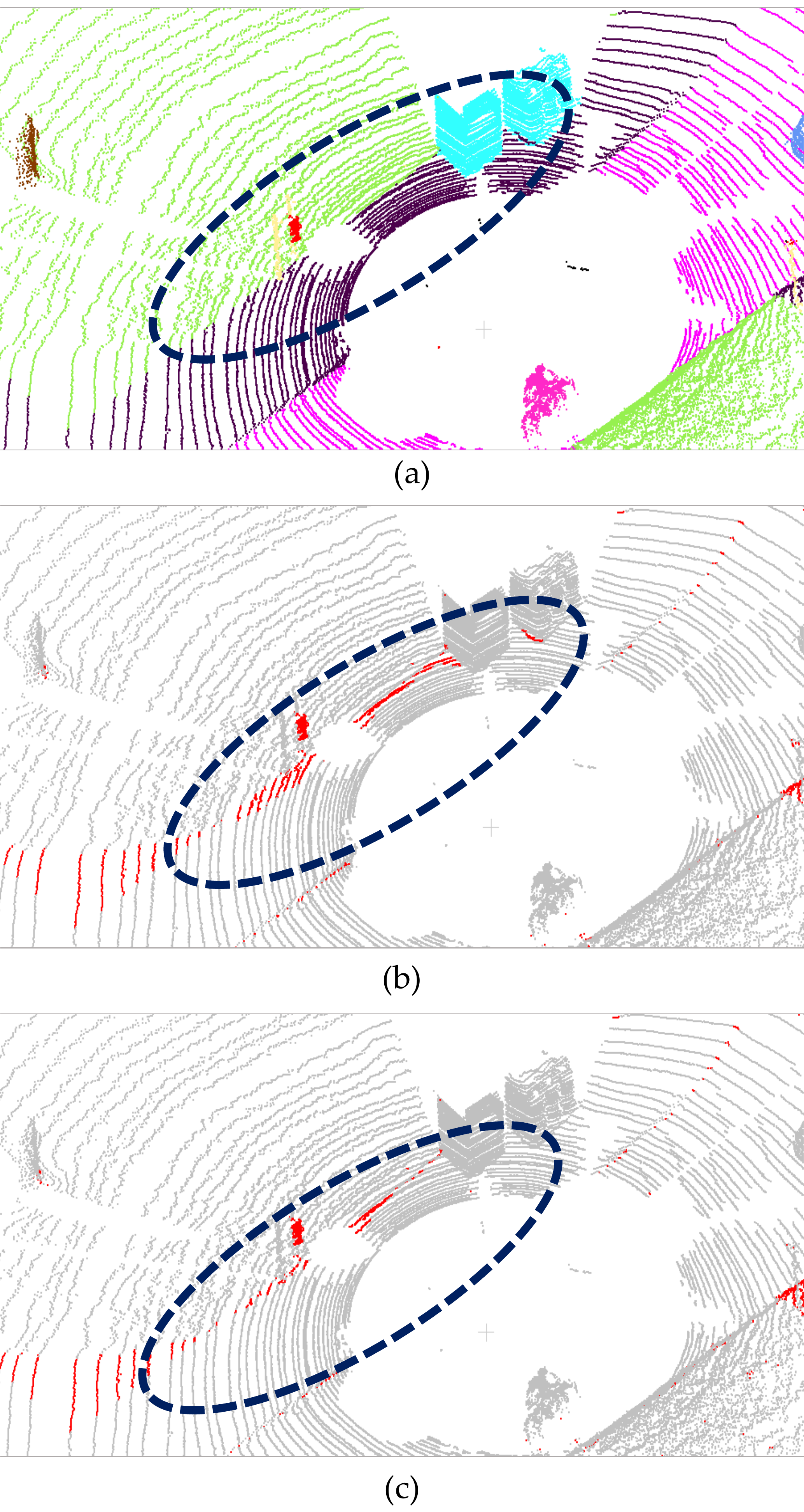}
\caption{Visualization of the combined loss effect. (a) Ground truth (b) Error of the model trained with cross-entropy loss (c) Error of the model trained with cross-entropy and position-aware(PA) loss. The error point between ground truth and prediction is red.}
\label{fig:combinedLoss}
\end{figure}

Also, Fig. \ref{fig:combinedLoss} shows the qualitative result of the experiment. The first column is the ground truth, and the next is the prediction of the model which is trained with only cross-entropy loss. The last image is the prediction of the combined loss trained model. As shown in the figure, the boundary points between sidewalk(purple) and vegetation(light green) are more well-segmented in the last model, which is trained with the combined loss. As a result, the proposed loss function contributes to the improvement in overall semantic segmentation performance, especially at the boundary points.

\begin{table}[!h]
\centering
\scriptsize
{
\begin{tabular}{|c|c|}
 \hline
 \textbf{Loss function} &\textbf{mIoU (\%)}\\
 \hline\hline
 Cross-entropy loss &  63.8  \\
 \hline
 \textbf{Cross-entropy + PA loss} &  \textbf{64.6} \\
 \hline
\end{tabular}
\caption{The mIoU of two models: Trained with only cross-entropy loss (upper),  Trained with cross-entropy and PA combined loss (lower). The mIoU measures in SemanticKITTI validation set.}
\label{table:combinedLoss}
}
\end{table}

\section{Conclusion}
\label{section:Conclusion}
In this paper, we proposed PCSCNet the fast voxel-based semantic segmentation model using point convolution and 3D sparse convolution. The proposed model was designed to target the real-time semantic segmentation of low voxel-resolution. For the purpose, we extract features using point convolution and propagates the features fastly using 3D sparse convolution. Then, we train the model using the cross-entropy loss and position-aware loss to leverage the discretization error from the low voxel-resolution. Finally, we conducted the extensive evaluation and discussion of the proposed model using SemanticKITTI and nuScenes. The proposed model achieved the state-of-the-art voxel-based method in real-time semantic segmentation, and also validated its performance in low voxel-resolution.




\bibliographystyle{cas-model2-names}

\bibliography{reference}



\end{document}